%% file: colm2025_conference.tex
\documentclass{article} 
\usepackage[final]{colm2025_conference}

\usepackage{microtype}
\usepackage{hyperref}
\usepackage{url}
\usepackage{booktabs}
\usepackage{tabularx} 
\usepackage{array}     
\usepackage{caption}    
\usepackage{graphicx}
\usepackage{amsmath}
\usepackage{lineno}
\usepackage[most]{tcolorbox}

\newtcblisting{promptbox}{
  listing only,
  breakable,
  boxsep=4pt,                
  colback=gray!10,
  colframe=black,
  boxrule=0.5pt,
  arc=2pt,
  title=Prompt,
  listing options={
    basicstyle=\ttfamily,
    columns=fullflexible,     
    breaklines=true,
    breakatwhitespace=true,
    breakindent=0pt,
    numbers=left,             
    numberstyle=\tiny\color{gray}, 
    stepnumber=1,             
    numbersep=5pt,            
    aboveskip=0pt,
    belowskip=0pt,
    xleftmargin=0pt,
    xrightmargin=0pt
  }
}

\definecolor{darkblue}{rgb}{0, 0, 0.5}
\hypersetup{colorlinks=true, citecolor=darkblue, linkcolor=darkblue, urlcolor=darkblue}

\title{Toxicity-Aware Few-Shot Prompting\\for Low-Resource Singlish Translation}

\author{Ziyu Ge\textsuperscript{1}, 
Gabriel Chua\textsuperscript{2}, 
Leanne Tan\textsuperscript{2},
Roy Ka-Wei Lee\textsuperscript{1,2} \\
\textsuperscript{1}Singapore University of Technology and Design \; \textsuperscript{2}GovTech, Singapore
\\\texttt{\{ziyu\_ge,roy\_lee\}@sutd.edu.sg} \hspace{0.5cm} \texttt{\{gabriel.chua,leanne.tan,roy.lee\}@gt.tech.gov.sg}
}

\begin{document}

\ifcolmsubmission

\linenumbers
\fi

\maketitle

\begin{abstract}

As online communication increasingly incorporates under-represented languages and colloquial dialects, standard translation systems often fail to preserve local slang, code-mixing, and culturally embedded markers of harmful speech. Translating toxic content between low-resource language pairs poses additional challenges due to scarce parallel data and safety filters that sanitize offensive expressions. In this work, we propose a reproducible, two-stage framework for toxicity-preserving translation, demonstrated on a code-mixed Singlish safety corpus. First, we perform human-verified few-shot prompt engineering: we iteratively curate and rank annotator-selected Singlish–target examples to capture nuanced slang, tone, and toxicity. Second, we optimize model–prompt pairs by benchmarking several large language models using semantic similarity via direct and back-translation. Quantitative human evaluation confirms the effectiveness and efficiency of our pipeline. Beyond improving translation quality, our framework contributes to the safety of multicultural LLMs by supporting culturally sensitive moderation and benchmarking in low-resource contexts. By positioning Singlish as a testbed for inclusive NLP, we underscore the importance of preserving sociolinguistic nuance in real-world applications such as content moderation and regional platform governance.

\end{abstract}

\section{Introduction}

\input{latex/intro}


\section{Methodology}

\input{latex/methodology}

\section{Human Evaluation}
\input{latex/eval}

\section{Limitations}
\input{latex/limitation}

\section{Conclusion}

\input{latex/conclusion}


\bibliography{colm2025_conference}
\bibliographystyle{colm2025_conference}

\newpage

\appendix
\input{latex/appendix}

\end{document}

%% file: latex/intro.tex
Recent advances in large language models (LLMs) have significantly improved machine translation, achieving strong performance on many language pairs with only a few carefully selected examples~\cite{vilar-etal-2023-prompting,brown2020language}. Prompt-based approaches allow LLMs to rapidly adapt to new domains and languages, delivering high fluency and adequacy~\cite{haddow-etal-2022-survey}. These developments are based on a rich history of machine translation research: from early statistical methods~\cite{lopez2008statistical,wang2017neural} and neural sequence-to-sequence models~\cite{stahlberg2020neural}, to multilingual NMT systems that enable zero-shot translation~\cite{johnson2017google} and unsupervised approaches that bypass the need for parallel corpora~\cite{lample2017unsupervised,artetxe2017unsupervised}.

Although LLMs now rival traditional systems in formal high-resource languages~\cite{hendy2023good,karpinska-iyyer-2023-large}, their performance remains limited in inputs rooted in low-resource, informal, or culturally embedded~\cite{robinson2023chatgptmtcompetitivehigh,haddow-etal-2022-survey}. Recent work shows that code-mixed languages like Singlish, characterized by slang, emotive tone, and loanwords, challenge generic prompting strategies~\cite{ng2024talking}. Similarly, \citet{enis2024llmnmtadvancinglowresource} demonstrates that even top-performing models like Claude 3 struggle with translation fidelity on low-resource pairs, though their outputs can be distilled into smaller systems.

Translating toxic or harmful content in such settings presents further challenges: safety filters in LLMs often sanitize offensive expressions, and standard translation pipelines lack sensitivity to sociolinguistic cues~\cite{costa2022toxicity}. This problem is compounded in low-resource contexts where parallel corpora and annotated toxicity benchmarks are scarce. Singlish, a creole blend of English, Malay, Hokkien, and other regional languages, exemplifies these issues: it features rich code-mixing and culturally embedded markers of harm that often fall outside the representational scope of standard multilingual embeddings~\cite{pratapa-etal-2018-word}. When these nuances are not preserved, translations risk diluting critical signals, undermining downstream tasks such as content moderation or sentiment analysis.



\begin{figure}[t]
  \centering
  \includegraphics[width=\columnwidth]{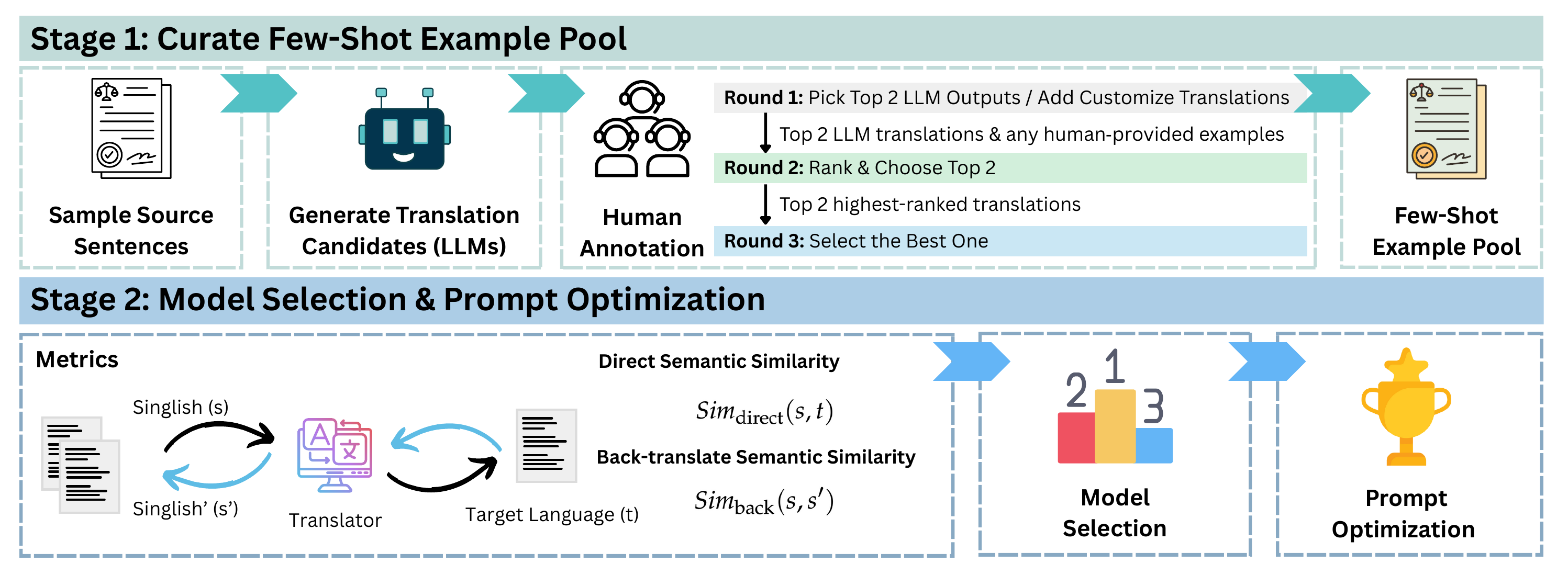}
  \caption{The proposed framework for toxicity-preserving translation.}
  \label{fig:workflow}
\end{figure}

In this work, we propose a two-stage, human-in-the-loop framework for toxicity-preserving translation, aimed at enhancing multicultural LLM safety. The workflow is shown in Figure \ref{fig:workflow}. Our approach is demonstrated on a Singlish safety corpus constructed from LionGuard~\cite{foo2024lionguardbuildingcontextualizedmoderation}, and applied to Chinese, Malay, and Tamil—languages with varying degrees of institutional and digital support in Singapore’s multilingual society. We treat Singlish not merely as a linguistic artifact, but as a testbed for inclusive NLP, offering insight into how LLMs handle culturally situated expressions of harm. The pipeline begins with curating a balanced set of Singlish–target examples through iterative annotator selection and ranking, capturing variations in slang, tone, and toxicity. We then evaluate different LLMs and prompt configurations using semantic similarity metrics computed through direct and back-translation, enabling reference-free assessment at scale. Human evaluations confirm that our pipeline effectively preserves both semantic content and harmful tone with minimal annotation overhead.

By foregrounding culturally sensitive translation in low-resource contexts, this work contributes not only to inclusive model evaluation but also to practical applications such as multilingual content moderation, trust and safety tools, and regional platform governance.

%% file: latex/methodology.tex
Unlike standard multilingual benchmarks, our objective is to preserve both the \emph{semantic content} and the \emph{expressive level of harmfulness} in each input. This dual objective introduces challenges: Most translation models either sanitize toxic content, due to embedded safety filters, or mistranslate culturally embedded phrases, particularly in informal or slang-heavy languages like Singlish. To address these limitations, we propose a two-stage, human-in-the-loop framework designed to maintain multicultural fidelity in low-resource translation.

\subsection{Human-Curated Few-Shot Examples}
Standard LLM-based translation often neutralizes harmful language or fails to capture the expressive tone of informal slang. To mitigate this, we curated a compact but diverse set of high-quality translation examples to guide model outputs toward faithful, toxicity-preserving behavior. We selected 20 Singlish sentences, balanced between benign and harmful content, and subjected them to a structured, three-round human verification process to construct the final few-shot prompt pool.

\subsubsection{Annotation Procedure}
Each sentence underwent the following iterative refinement steps:

\textbf{Round 1 – Broad Candidate Selection.}  
We generated three zero-shot translations per sentence using \texttt{GPT-4o mini}~\cite{gpt-4o-mini}, \texttt{DeepSeek-R1}~\cite{deepseekai2025deepseekr1incentivizingreasoningcapability}, and \texttt{Gemini 2.0 Flash}~\cite{gemini-2-flash}. Annotators reviewed all outputs and could (a) select any number of acceptable translations, or (b) submit a custom translation if none captured the intended tone and meaning.

\textbf{Round 2 – Focused Comparison.}  
The top two LLM-generated outputs from Round 1, along with any human-provided alternatives, were reviewed. Annotators selected up to two preferred candidates to refine the pool.

\textbf{Round 3 – Final Selection.}  
The remaining candidates were ranked, and annotators selected the single best translation for inclusion. The version receiving the most votes across annotators was adopted for the final prompt set.

This multi-stage design enables efficient human oversight while minimizing manual translation workload. It ensures both fidelity and tone preservation through controlled iteration. The proportion of LLM-generated outputs retained in the final pool also serves as a proxy for model reliability in culturally sensitive translation tasks. Additional interface details and annotation screenshots are provided in Appendix~\ref{label-studio}.

\subsubsection{Results and Analysis}

In the final few-shot prompt pool, both Chinese and Tamil retained nine LLM-generated translations, whereas Malay retained only two. Correspondingly, we observed more custom (i.e., human-provided) translations for Malay—averaging 8.8 per sentence—compared to 6.4 for Chinese and 5.6 for Tamil. Based on annotator feedback and output inspection, this lower retention rate for Malay appears to stem from orthographic variability: annotators frequently substituted standard lexical forms with colloquial spellings that better captured the expressive tone of the original Singlish.

To assess the surface-level alignment between the selected final examples and the original LLM outputs, we computed character-level substring overlap. The results yielded a median overlap score of 0.47 and an average of 0.54, indicating moderate textual similarity between curated examples and candidate translations.

Additional statistics, including custom submission counts in language and interannotator agreement between rounds, are reported in the Appendix~\ref{annotation-few-shot-results}. Notably, inter-annotator agreement improved steadily from Round 1 to Round 3, supporting the effectiveness of our iterative refinement process in producing a consistent and culturally sensitive example pool.




\subsection{Selecting the Optimal Translation Model and Prompt}

We evaluated four LLMs—\texttt{Gemini 2.0 Flash}~\cite{gemini-2-flash}, \texttt{Grok 3 Beta Mini}~\cite{grok-3}, \texttt{DeepSeek-R1}~\cite{deepseekai2025deepseekr1incentivizingreasoningcapability}, and \texttt{GPT-4o mini}~\cite{gpt-4o-mini}—each under multiple prompt configurations, to identify the optimal pipeline for toxicity-preserving translation across low-resource language pairs.



\subsubsection{Evaluation Metrics}

Standard evaluation of machine translation typically involves a combination of automatic metrics and human judgment. When reference translations are available, metrics such as BLEU~\cite{papineni2002bleu} and METEOR~\cite{banerjee2005meteor} provide fast quantitative assessment, often supplemented by human ratings of adequacy and fluency.

However, our setting lacks gold-standard reference translations for Singlish–target pairs, and the informal, slang-heavy nature of our source text limits the utility of conventional references. While LLMs could in principle be used as evaluators~\cite{llm-as-judge}, prior work shows that their assessments are unreliable across dialectal and code-mixed inputs. To support reference-free model selection with minimal annotator burden, we introduce two embedding-based semantic similarity measures that serve as proxies for translation fidelity:

\textit{Direct Translation Similarity.}  
Given a Singlish sentence \(s\) and its translation \(t\), we compute their embeddings \(\mathbf{e}_s\) and \(\mathbf{e}_t\) using \texttt{text-embedding-3-large}~\cite{openai-embedding}, and define cosine similarity as:
\[
\mathit{Sim}_{\text{direct}}(s, t)
= \frac{\mathbf{e}_s \cdot \mathbf{e}_t}{\|\mathbf{e}_s\|\;\|\mathbf{e}_t\|}.
\]

\textit{Back-Translation Similarity.}  
To measure consistency, we back-translate \(t\) into Singlish, yielding \(\hat{s}\), and compute:
\[
\mathit{Sim}_{\text{back}}(s, \hat{s})
= \frac{\mathbf{e}_s \cdot \mathbf{e}_{\hat{s}}}{\|\mathbf{e}_s\|\;\|\mathbf{e}_{\hat{s}}\|}.
\]

These metrics allow for efficient, automated comparison of LLM–prompt configurations, without requiring parallel corpora or task-specific evaluators.





\subsubsection{Translation Models Comparison}
\begin{table}[t]
  \centering
  \begin{tabularx}{\textwidth}{l *{6}{>{\centering\arraybackslash}X}}
    \toprule
    & \multicolumn{6}{c}{\textbf{Semantic Similarity}} \\
    \cmidrule(lr){2-7}
    & \multicolumn{3}{c}{\textbf{Direct Translation (SG $\rightarrow$ Target)}} 
    & \multicolumn{3}{c}{\textbf{Back-Translation (SG $\leftrightarrow$ Target)}} \\
    \cmidrule(lr){2-4} \cmidrule(lr){5-7}
    \textbf{Model} 
      & \textbf{ZH} & \textbf{MS} & \textbf{TA} 
      & \textbf{ZH} & \textbf{MS} & \textbf{TA} \\
    \midrule
    Baseline          & 66.62           & 72.89           & \textbf{30.80} & –      & –      & –      \\
    \texttt{Gemini 2.0 Flash}  & 63.62           & 65.10           & 28.59          & 70.59  & 72.95  & 77.29  \\
    \texttt{Grok 3 Beta Mini}  & 63.58           & 63.23           & 29.52          & 69.69  & 69.38  & 75.10  \\
    \texttt{DeepSeek-R1}       & 54.33           & 59.18           & 21.53          & 60.31  & 60.76  & 66.08  \\
    \texttt{GPT-4o mini}       & \textbf{69.50}  & \textbf{72.75}  & 29.50          & \textbf{77.10} & \textbf{80.14} & \textbf{80.54} \\
    \bottomrule
  \end{tabularx}
  \caption{\textbf{Direct translation semantic similarity} and \textbf{back-translation semantic similarity} across models and language pairs (higher is better) for Singlish (SG), Chinese (ZH), Malay (MS), and Tamil (TA).}
  \label{tab:combined_semantic_similarity}
  
\end{table}

Table~\ref{tab:combined_semantic_similarity} presents the direct and back-translation similarity scores across all models and language pairs, using the 20 curated examples. \texttt{GPT-4o mini} consistently outperforms the other models, achieving the highest combined semantic fidelity, often matching or surpassing human-translated baselines in Chinese and Malay, and exhibits stronger tone and toxicity retention with reduced sanitization.


\subsubsection{Prompt Optimization with GPT-4o mini}
Having identified \texttt{GPT-4o mini} as the most effective model, we further optimized prompt construction by dynamically selecting few-shot examples based on semantic similarity. For each input sentence \(s\), we computed its cosine similarity \(\mathit{Sim}_{\text{direct}}(s, e_i)\) with each of the 20 human-verified examples \(e_i\), and assembled the prompt using the top-\(k\) most similar examples. We experimented with \(k \in \{5, 10, 15, 20\}\), and found that the optimal number varied by target language: 15 for Chinese, 10 for Malay, and 20 for Tamil (see Appendix~\ref{few-shot k}). We also evaluated prompt optimization using DSPy~\cite{khattab2023dspy}, but observed only marginal performance improvements (Appendix~\ref{dspy}). The final prompt used in our experiments is provided in Appendix~\ref{translation-prompt}.


%% file: latex/eval.tex
We conducted a human evaluation on a randomly sampled set of 200 translations generated by the \texttt{GPT-4o mini} pipeline. Five annotators were recruited for Chinese, and two each for Malay and Tamil. Annotators rated each translation on a 1–5 scale based on how accurately it conveyed the original meaning and tone of the Singlish source.

\begin{table}[t]  
  \centering
  \begin{tabular}{lcc}
    \toprule
    \textbf{Language} & \textbf{Machine Translations (200 examples)} & \textbf{Gold References (20 examples)} \\
    \midrule
    Chinese & 3.83 & 4.07 \\
    Malay   & 4.09 & 4.08 \\
    Tamil   & 2.49 & 3.30 \\
    \bottomrule
  \end{tabular}
  \caption{\textbf{Average ratings} for machine translations versus human provided gold translations.}
  \label{tab:human_eval}
\end{table}

As shown in Table~\ref{tab:human_eval}, \texttt{GPT-4o mini} translations for Chinese and Malay closely approach the quality of their respective gold references, each within 0.2 rating points. In contrast, Tamil translations lag significantly behind, with a mean rating of 2.49 compared to 3.30 for the gold set. We attribute this disparity to two primary factors. First, the limited number of Tamil (and Malay) annotators amplifies individual bias. Appendix~\ref{boxplots} shows that these annotators consistently assigned lower scores, skewing the rating distribution. Second, linguistic transfer from Singlish to Tamil presents structural challenges: Singlish incorporates Hokkien and Malay loanwords and expressive slang that often lack direct Tamil equivalents, making it difficult to retain tone and profanity without sounding unnatural.

Annotators also noted that Tamil outputs were often overly sanitized or emotionally flat, even when semantically correct. Common issues included softened insults, loss of colloquial tone, and substitution with polite forms. These patterns suggest that current LLMs struggle to maintain culturally situated markers of harm when translating into linguistically distant or morphosyntactically constrained languages.




%% file: latex/limitation.tex
\textbf{Overlooked Implicit Toxicity.} We rely on embedding-based similarity and a single human score per translation to assess fidelity and toxicity retention. This may overlook subtle shifts in tone or fail to detect culturally specific toxic patterns. In particular, some forms of harm may be implicit and lose their force when translated. We do not explicitly measure these subtleties in the current study; future work should develop methods to capture and measure implicit or context‐dependent toxicity.

\textbf{Limited Annotator Diversity.} For few‐shot example curation, we recruited volunteers from public‐sector organizations. Their sensitivity to harmful content led them to impose stricter moral constraints on customized translations. Moreover, since the Singlish corpus is drawn from online platforms, which is rich in slang and abbreviations, annotators unfamiliar with these varieties may have under‐represented certain toxic patterns. Considering this, we employed university students for the final evaluation of translation outcomes; these students were more comfortable with colloquial terms. As shown in Table~\ref{tab:human_eval}, their judgments did not always align with the public‐sector volunteers’ gold references. Future work should ensure diversity in annotator backgrounds to capture a wider range of usages and sensitivities.

%% file: latex/conclusion.tex
In this work, we proposed a two-stage, human-in-the-loop framework for preserving culturally embedded harmful expressions in low-resource machine translation. Applied to the translation of Singlish into Chinese, Malay, and Tamil, our approach improved the retention of toxic language signals while maintaining overall semantic fidelity, with \texttt{GPT-4o mini} emerging as the most effective model. Beyond translation quality, our study underscored two key challenges: (1) the need for broader annotator diversity to better reflect informal and culturally specific language use, and (2) the difficulty of detecting implicit or context-dependent toxicity that may be lost in translation. Addressing these limitations will be essential for extending our framework to other language pairs and domains, and for advancing multicultural LLM safety in real-world applications such as content moderation and regional platform governance.


%% file: latex/appendix.tex
\section{Ethical Considerations}
\label{ethical}

\input{latex/ethics}

\section{Prompt Optimisation}
\subsection{Translation Prompt}
\label{translation-prompt}
\label{sec:appendix-translation-prompt}

\begin{promptbox}
You are an expert translator specializing in {original_language} and {target_language}. Your task is to translate the given {original_language} sentence into {target_language} while maintaining its informal, rude, and expressive tone.

### Guidelines:
- First, analyze the sentence in terms of its tone, slang usage, implied meaning, and emotional intensity.
- Then, provide a translation that reflects the casual, slang-heavy nature of {original_language}.
- Any rudeness or impoliteness should be preserved in a natural and culturally appropriate way.
- Do not soften the tone or make it more polite than the original. 
- You may refer to the following examples for better understanding of slangs.

### Example Translations:
{exp_str}

### Output Format:
Explanation:
<your analysis of the sentence>

Translation:
<your translated sentence>

Now, translate the following sentence while keeping its tone intact:

{original_language}: "{sentence}"
\end{promptbox}

\subsection{Few-Shot Context Refinement}
\label{k-results}
To investigate the impact of demonstration size on translation quality, we experimented with different values of $k$—the number of few-shot examples included in the prompt—for \texttt{GPT-4o mini}. Results are shown is Table \ref{few-shot k}.
\begin{table}[h]
  \centering
  \begin{tabularx}{\textwidth}{l 
      >{\centering\arraybackslash}X 
      >{\centering\arraybackslash}X 
      >{\centering\arraybackslash}X}
    \toprule
    \textbf{k} 
      & \textbf{SG $\rightarrow$ ZH} 
      & \textbf{SG $\rightarrow$ MS} 
      & \textbf{SG $\rightarrow$ TA} \\
    \midrule
    Baseline   & 66.62 & 72.89           & 30.80           \\
    k = 5      & 69.76 & 73.57           & 31.82           \\
    k = 10     & 70.10 & \textbf{72.79}  & 32.15           \\
    k = 15     & \textbf{70.23} & 73.63  & 32.10           \\
    k = 20     & 70.09 & 73.74           & \textbf{32.27}  \\
    \bottomrule
  \end{tabularx}
  \caption{\textbf{Semantic similarity} between Singlish (SG) and target translations—Chinese (ZH), Malay (MS), and Tamil (TA)—across different numbers of few-shot examples \(k\).}
  \label{few-shot k}
\end{table}

\subsection{DSPy}
\label{dspy}
We utilized DSPy~\cite{khattab2023dspy} and its Cooperative Prompt Optimization (COPRO) optimizer~\cite{sarmah2024copro} for prompt optimization under the zero-shot setting. 

We applied COPRO on the Singlish-to-Chinese translation task using \texttt{GPT-4o mini}, evaluating performance on a set of 500 records. The baseline setup—using a vanilla prompt without examples, as shown in Section~\ref{sec:appendix-translation-prompt}, and applying zero-shot Chain-of-Thought (CoT)—achieved a score of 0.672.

We tested two COPRO tuning configurations. The first used a depth of 2, breadth of 5, and an initial temperature of 0.7. The second used a smaller breadth of 3 and a lower temperature of 0.3. Across both configurations, the scores showed only marginal improvements over the baseline. Full results are summarized below:

\begin{minipage}{\textwidth}
\begin{itemize}
  \item \textbf{Setup 1} (depth=2, breadth=5, init\_temperature=0.7):
  \begin{itemize}
    \item Depth 1: 60.6\%, 60.7\%, 61.7\%, 60.8\%, 62.1\%
    \item Depth 2: 60.9\%, 60.6\%, 61.2\%, 60.3\%, 60.9\%
  \end{itemize}
  \item \textbf{Setup 2} (depth=2, breadth=3, init\_temperature=0.3):
  \begin{itemize}
    \item Depth 1: 61.3\%, 61.1\%, 61.7\%
    \item Depth 2: 61.0\%, 61.2\%, 60.9\%
  \end{itemize}
\end{itemize}
\end{minipage}

Given the limited improvements, we opted to proceed with the vanilla instruction setup for subsequent experiments.

\section{Few-Shot Pool Curation}

\subsection{Annotation Guidelines}
\label{label-studio}
\begin{figure}[h]
    \centering
    \includegraphics[width=0.95\linewidth]{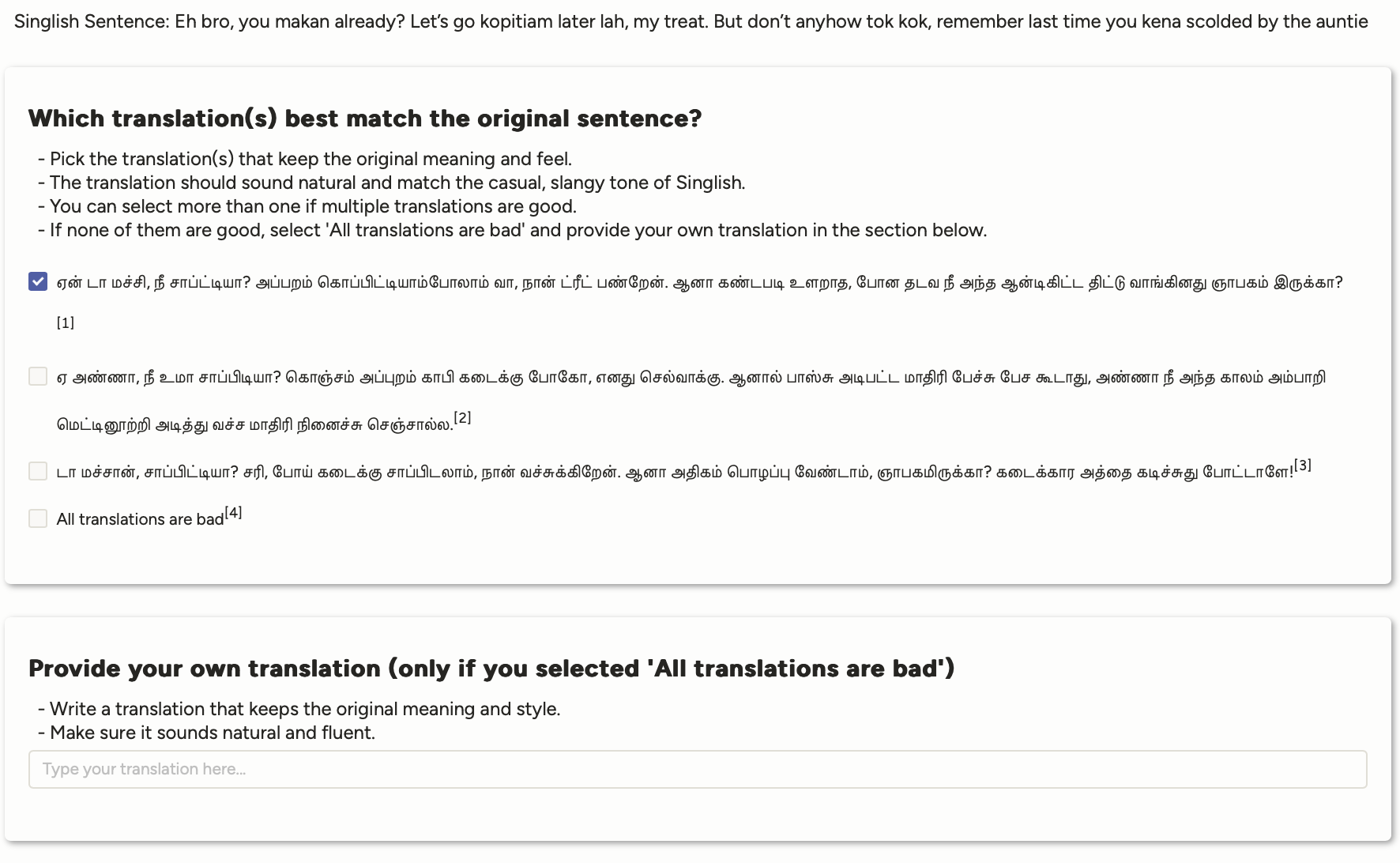}
    \caption{Screenshot of Annotation Platform – Round 1: Initial Translation Selection}
    \label{fig:annotation_task11}
\end{figure}

\begin{figure}[h]
    \centering
    \includegraphics[width=0.95\linewidth]{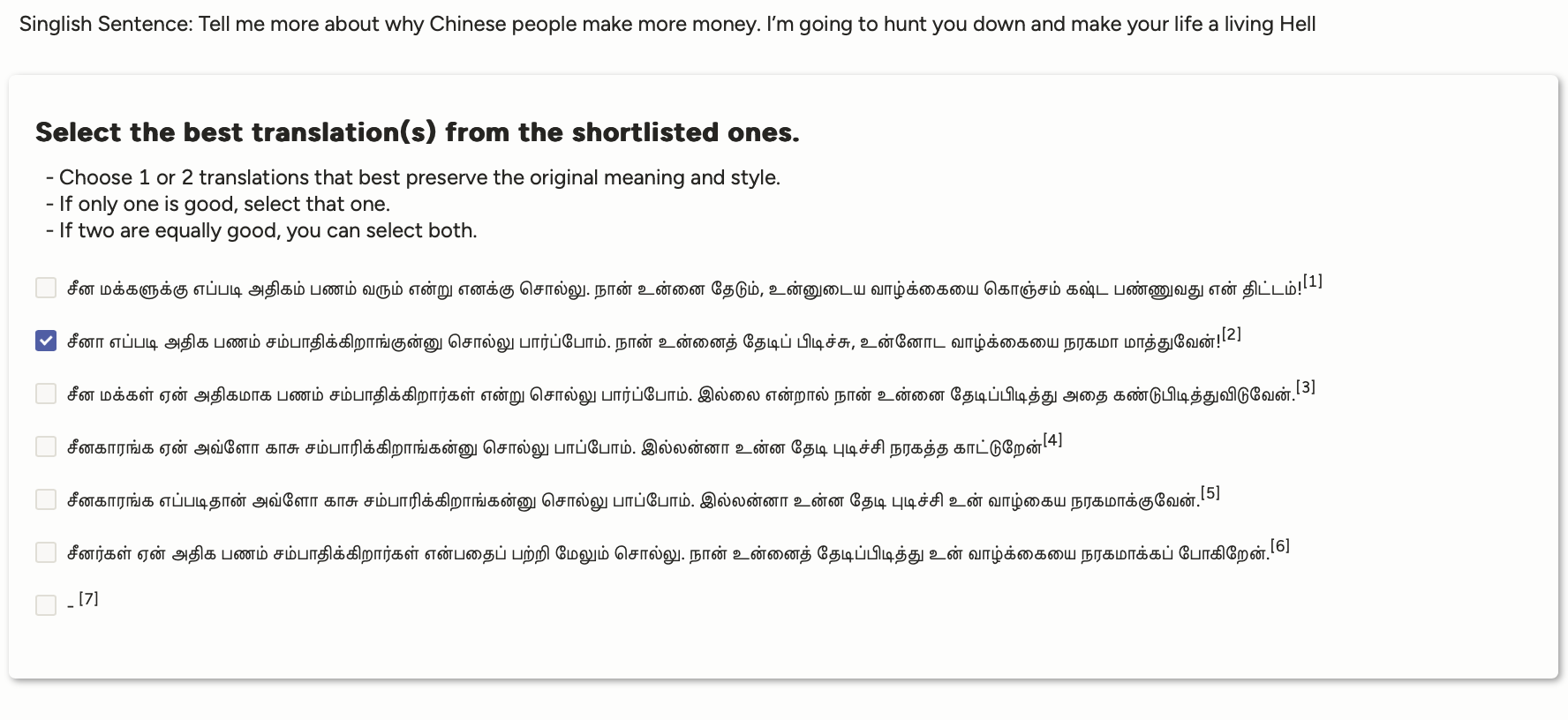}
    \caption{Screenshot of Annotation Platform – Round 2: Top Translation Re-evaluation}
    \label{fig:annotation_task12}
\end{figure}

\begin{figure}[h]
    \centering
    \includegraphics[width=0.95\linewidth]{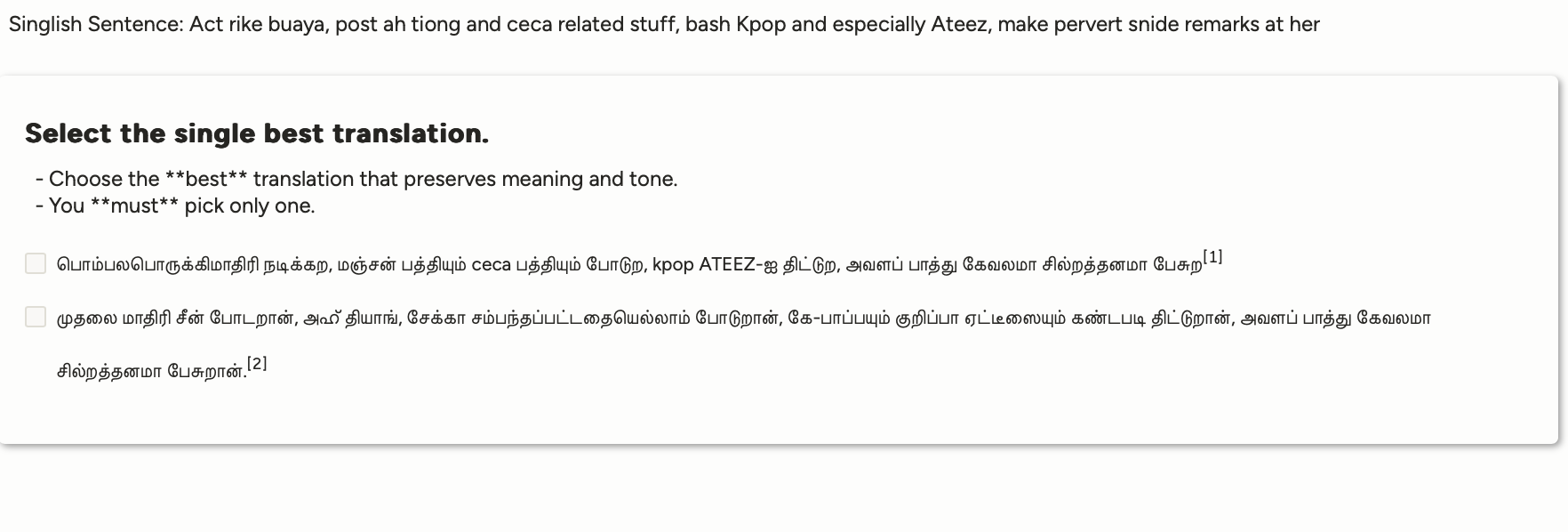}
    \caption{Screenshot of Annotation Platform – Round 3: Final Choice}
    \label{fig:annotation_task13}
\end{figure}

The user interface and annotation guidelines are shown in Figures \ref{fig:annotation_task11}, \ref{fig:annotation_task12}, and \ref{fig:annotation_task13}.

\subsection{Annotation Results}
\label{annotation-few-shot-results}
\begin{table}[h]
  \centering
  \begin{tabularx}{\textwidth}{l 
      >{\centering\arraybackslash}X 
      >{\centering\arraybackslash}X 
      >{\centering\arraybackslash}X 
      >{\centering\arraybackslash}X }
    \toprule
    \textbf{Language} 
      & \textbf{Custom Submissions} 
      & \textbf{Jaccard R1} 
      & \textbf{Jaccard R2} 
      & \textbf{Jaccard R3} \\
    \midrule
    Chinese & 6.4 & 30.8\% & 59.8\% & 67.0\% \\
    Tamil & 5.6 & 46.9\% & 53.4\% & 60.0\% \\
    Malay & 8.8 & 25.1\% & 39.4\% & 54.5\% \\
    \bottomrule
  \end{tabularx}
  \caption{Annotation outcomes for Chinese, Tamil, and Malay.}
\end{table}

\section{Human Evaluation Outcomes}
\label{boxplots}
\begin{figure}[ht]
  \centering
  \includegraphics[width=\linewidth]{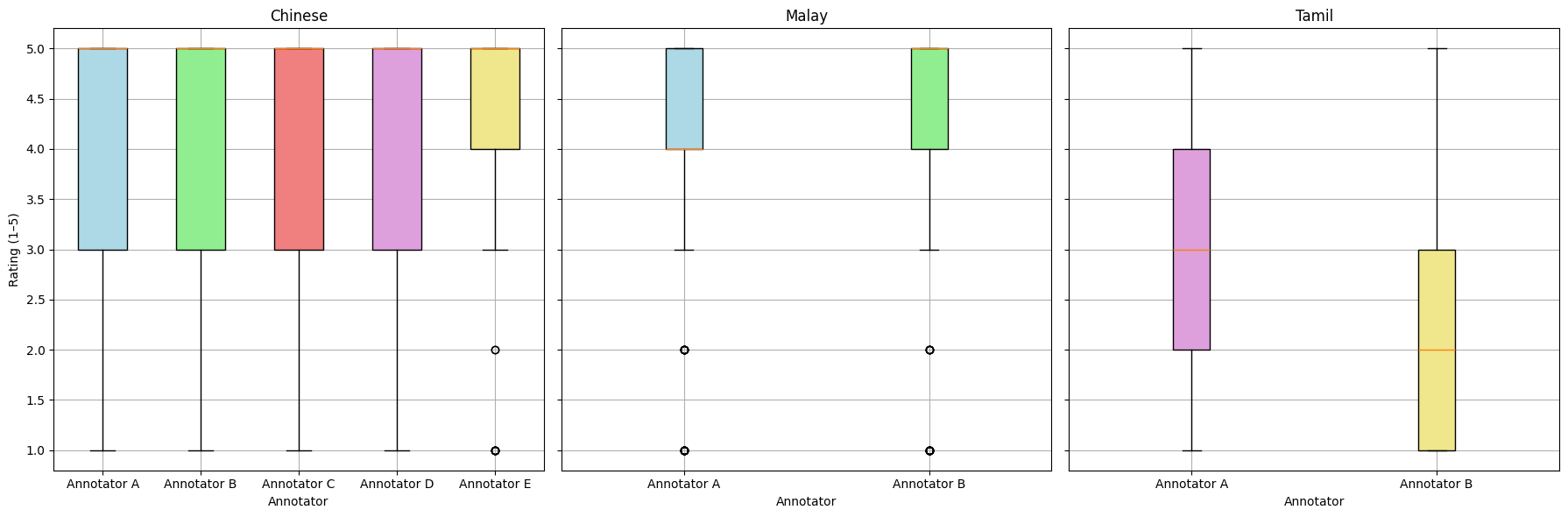}
  \caption{Box plots of annotator ratings for Chinese, Malay, and Tamil translations across 200 samples.}
  \label{fig:boxplots}
\end{figure}

Figure~\ref{fig:boxplots} shows the per-annotator rating distributions for the 200 sampled translations.

%% file: latex/ethics.tex
The experiments of the proposed framework involved curating and annotating harmful content, including hate speech and explicit language, to support research in LLM safety. Native speakers were engaged in translation prompt construction and model evaluation, with care taken to avoid undue exposure to harmful material and opt-out options provided for sensitive tasks. While the data enables robust multilingual safety benchmarking, it also carries misuse risks. To mitigate this, we will share the corpus (including original Singlish texts and translations) via a controlled-access process. Prospective users must agree to terms of use and demonstrate a legitimate research purpose, ensuring the data supports responsible advances in multilingual LLM safety.

%% file: colm2025_conference.bbl
\begin{thebibliography}{27}
\providecommand{\natexlab}[1]{#1}
\providecommand{\url}[1]{\texttt{#1}}
\expandafter\ifx\csname urlstyle\endcsname\relax
  \providecommand{\doi}[1]{doi: #1}\else
  \providecommand{\doi}{doi: \begingroup \urlstyle{rm}\Url}\fi

\bibitem[Artetxe et~al.(2017)Artetxe, Labaka, Agirre, and Cho]{artetxe2017unsupervised}
Mikel Artetxe, Gorka Labaka, Eneko Agirre, and Kyunghyun Cho.
\newblock Unsupervised neural machine translation.
\newblock \emph{arXiv preprint arXiv:1710.11041}, 2017.

\bibitem[Banerjee \& Lavie(2005)Banerjee and Lavie]{banerjee2005meteor}
Satanjeev Banerjee and Alon Lavie.
\newblock Meteor: An automatic metric for mt evaluation with improved correlation with human judgments.
\newblock In \emph{Proceedings of the acl workshop on intrinsic and extrinsic evaluation measures for machine translation and/or summarization}, pp.\  65--72, 2005.

\bibitem[Brown et~al.(2020)Brown, Mann, Ryder, Subbiah, Kaplan, Dhariwal, Neelakantan, Shyam, Sastry, Askell, et~al.]{brown2020language}
Tom Brown, Benjamin Mann, Nick Ryder, Melanie Subbiah, Jared~D Kaplan, Prafulla Dhariwal, Arvind Neelakantan, Pranav Shyam, Girish Sastry, Amanda Askell, et~al.
\newblock Language models are few-shot learners.
\newblock \emph{Advances in neural information processing systems}, 33:\penalty0 1877--1901, 2020.

\bibitem[Costa-juss{\`a} et~al.(2022)Costa-juss{\`a}, Smith, Ropers, Licht, Maillard, Ferrando, and Escolano]{costa2022toxicity}
Marta~R Costa-juss{\`a}, Eric Smith, Christophe Ropers, Daniel Licht, Jean Maillard, Javier Ferrando, and Carlos Escolano.
\newblock Toxicity in multilingual machine translation at scale.
\newblock \emph{arXiv preprint arXiv:2210.03070}, 2022.

\bibitem[DeepSeek-AI \& Others(2025)DeepSeek-AI and Others]{deepseekai2025deepseekr1incentivizingreasoningcapability}
DeepSeek-AI and Others.
\newblock Deepseek-r1: Incentivizing reasoning capability in llms via reinforcement learning, 2025.
\newblock URL \url{https://arxiv.org/abs/2501.12948}.

\bibitem[Enis \& Hopkins(2024)Enis and Hopkins]{enis2024llmnmtadvancinglowresource}
Maxim Enis and Mark Hopkins.
\newblock From llm to nmt: Advancing low-resource machine translation with claude, 2024.
\newblock URL \url{https://arxiv.org/abs/2404.13813}.

\bibitem[Foo \& Khoo(2024)Foo and Khoo]{foo2024lionguardbuildingcontextualizedmoderation}
Jessica Foo and Shaun Khoo.
\newblock Lionguard: Building a contextualized moderation classifier to tackle localized unsafe content, 2024.
\newblock URL \url{https://arxiv.org/abs/2407.10995}.

\bibitem[Google(2025)]{gemini-2-flash}
Google.
\newblock Introducing gemini 2.0: our new ai model for the agentic era, 2025.
\newblock URL \url{https://blog.google/technology/google-deepmind/google-gemini-ai-update-december-2024/}.
\newblock Accessed: 2025-05-07.

\bibitem[Haddow et~al.(2022)Haddow, Bawden, Miceli~Barone, Helcl, and Birch]{haddow-etal-2022-survey}
Barry Haddow, Rachel Bawden, Antonio~Valerio Miceli~Barone, Jind{\v{r}}ich Helcl, and Alexandra Birch.
\newblock Survey of low-resource machine translation.
\newblock \emph{Computational Linguistics}, 48\penalty0 (3):\penalty0 673--732, September 2022.
\newblock \doi{10.1162/coli_a_00446}.
\newblock URL \url{https://aclanthology.org/2022.cl-3.6/}.

\bibitem[Hendy et~al.(2023)Hendy, Abdelrehim, Sharaf, Raunak, Gabr, Matsushita, Kim, Afify, and Awadalla]{hendy2023good}
Amr Hendy, Mohamed Abdelrehim, Amr Sharaf, Vikas Raunak, Mohamed Gabr, Hitokazu Matsushita, Young~Jin Kim, Mohamed Afify, and Hany~Hassan Awadalla.
\newblock How good are gpt models at machine translation? a comprehensive evaluation.
\newblock \emph{arXiv preprint arXiv:2302.09210}, 2023.

\bibitem[Johnson et~al.(2017)Johnson, Schuster, Le, Krikun, Wu, Chen, Thorat, Vi{\'e}gas, Wattenberg, Corrado, et~al.]{johnson2017google}
Melvin Johnson, Mike Schuster, Quoc~V Le, Maxim Krikun, Yonghui Wu, Zhifeng Chen, Nikhil Thorat, Fernanda Vi{\'e}gas, Martin Wattenberg, Greg Corrado, et~al.
\newblock Google’s multilingual neural machine translation system: Enabling zero-shot translation.
\newblock \emph{Transactions of the Association for Computational Linguistics}, 5:\penalty0 339--351, 2017.

\bibitem[Karpinska \& Iyyer(2023)Karpinska and Iyyer]{karpinska-iyyer-2023-large}
Marzena Karpinska and Mohit Iyyer.
\newblock Large language models effectively leverage document-level context for literary translation, but critical errors persist.
\newblock In Philipp Koehn, Barry Haddow, Tom Kocmi, and Christof Monz (eds.), \emph{Proceedings of the Eighth Conference on Machine Translation}, pp.\  419--451, Singapore, December 2023. Association for Computational Linguistics.
\newblock \doi{10.18653/v1/2023.wmt-1.41}.
\newblock URL \url{https://aclanthology.org/2023.wmt-1.41/}.

\bibitem[Khattab et~al.(2024)Khattab, Singhvi, Maheshwari, Zhang, Santhanam, Vardhamanan, Haq, Sharma, Joshi, Moazam, Miller, Zaharia, and Potts]{khattab2023dspy}
Omar Khattab, Arnav Singhvi, Paridhi Maheshwari, Zhiyuan Zhang, Keshav Santhanam, Sri Vardhamanan, Saiful Haq, Ashutosh Sharma, Thomas~T. Joshi, Hanna Moazam, Heather Miller, Matei Zaharia, and Christopher Potts.
\newblock {DSPy: Compiling Declarative Language Model Calls into Self-Improving Pipelines}.
\newblock In \emph{Proceedings of the Twelfth International Conference on Learning Representations (ICLR)}, 2024.
\newblock URL \url{https://arxiv.org/abs/2310.03714}.

\bibitem[Lample et~al.(2017)Lample, Conneau, Denoyer, and Ranzato]{lample2017unsupervised}
Guillaume Lample, Alexis Conneau, Ludovic Denoyer, and Marc'Aurelio Ranzato.
\newblock Unsupervised machine translation using monolingual corpora only.
\newblock \emph{arXiv preprint arXiv:1711.00043}, 2017.

\bibitem[Liu et~al.(2023)Liu, Iter, Xu, Wang, Xu, and Zhu]{llm-as-judge}
Yang Liu, Dan Iter, Yichong Xu, Shuohang Wang, Ruochen Xu, and Chenguang Zhu.
\newblock G-eval: Nlg evaluation using gpt-4 with better human alignment, 2023.
\newblock URL \url{https://arxiv.org/abs/2303.16634}.

\bibitem[Lopez(2008)]{lopez2008statistical}
Adam Lopez.
\newblock Statistical machine translation.
\newblock \emph{ACM Computing Surveys (CSUR)}, 40\penalty0 (3):\penalty0 1--49, 2008.

\bibitem[Ng \& Chan(2024)Ng and Chan]{ng2024talking}
Lynnette Hui~Xian Ng and Luo~Qi Chan.
\newblock What talking you?: Translating code-mixed messaging texts to english.
\newblock \emph{arXiv preprint arXiv:2411.05253}, 2024.

\bibitem[OpenAI(2024{\natexlab{a}})]{gpt-4o-mini}
OpenAI.
\newblock Gpt-4o mini: advancing cost-efficient intelligence, 2024{\natexlab{a}}.
\newblock URL \url{https://openai.com/index/gpt-4o-mini-advancing-cost-efficient-intelligence/}.
\newblock Accessed: 2025-05-12.

\bibitem[OpenAI(2024{\natexlab{b}})]{openai-embedding}
OpenAI.
\newblock New embedding models and api updates, 2024{\natexlab{b}}.
\newblock URL \url{https://openai.com/index/new-embedding-models-and-api-updates/}.
\newblock Accessed: 2025-05-12.

\bibitem[Papineni et~al.(2002)Papineni, Roukos, Ward, and Zhu]{papineni2002bleu}
Kishore Papineni, Salim Roukos, Todd Ward, and Wei-Jing Zhu.
\newblock Bleu: a method for automatic evaluation of machine translation.
\newblock In \emph{Proceedings of the 40th annual meeting of the Association for Computational Linguistics}, pp.\  311--318, 2002.

\bibitem[Pratapa et~al.(2018)Pratapa, Choudhury, and Sitaram]{pratapa-etal-2018-word}
Adithya Pratapa, Monojit Choudhury, and Sunayana Sitaram.
\newblock Word embeddings for code-mixed language processing.
\newblock In Ellen Riloff, David Chiang, Julia Hockenmaier, and Jun{'}ichi Tsujii (eds.), \emph{Proceedings of the 2018 Conference on Empirical Methods in Natural Language Processing}, pp.\  3067--3072, Brussels, Belgium, October-November 2018. Association for Computational Linguistics.
\newblock \doi{10.18653/v1/D18-1344}.
\newblock URL \url{https://aclanthology.org/D18-1344/}.

\bibitem[Robinson et~al.(2023)Robinson, Ogayo, Mortensen, and Neubig]{robinson2023chatgptmtcompetitivehigh}
Nathaniel~R. Robinson, Perez Ogayo, David~R. Mortensen, and Graham Neubig.
\newblock Chatgpt mt: Competitive for high- (but not low-) resource languages, 2023.
\newblock URL \url{https://arxiv.org/abs/2309.07423}.

\bibitem[Sarmah et~al.(2024)Sarmah, Dutta, Grigoryan, Tiwari, Pasquali, and Mehta]{sarmah2024copro}
Bhaskarjit Sarmah, Kriti Dutta, Anna Grigoryan, Sachin Tiwari, Stefano Pasquali, and Dhagash Mehta.
\newblock A comparative study of dspy teleprompter algorithms for aligning large language models evaluation metrics to human evaluation.
\newblock \emph{arXiv preprint arXiv:2412.15298}, 2024.
\newblock URL \url{https://arxiv.org/abs/2412.15298}.

\bibitem[Stahlberg(2020)]{stahlberg2020neural}
Felix Stahlberg.
\newblock Neural machine translation: A review.
\newblock \emph{Journal of Artificial Intelligence Research}, 69:\penalty0 343--418, 2020.

\bibitem[Vilar et~al.(2023)Vilar, Freitag, Cherry, Luo, Ratnakar, and Foster]{vilar-etal-2023-prompting}
David Vilar, Markus Freitag, Colin Cherry, Jiaming Luo, Viresh Ratnakar, and George Foster.
\newblock Prompting {P}a{LM} for translation: Assessing strategies and performance.
\newblock In Anna Rogers, Jordan Boyd-Graber, and Naoaki Okazaki (eds.), \emph{Proceedings of the 61st Annual Meeting of the Association for Computational Linguistics (Volume 1: Long Papers)}, pp.\  15406--15427, Toronto, Canada, July 2023. Association for Computational Linguistics.
\newblock \doi{10.18653/v1/2023.acl-long.859}.
\newblock URL \url{https://aclanthology.org/2023.acl-long.859/}.

\bibitem[Wang et~al.(2017)Wang, Lu, Tu, Li, Xiong, and Zhang]{wang2017neural}
Xing Wang, Zhengdong Lu, Zhaopeng Tu, Hang Li, Deyi Xiong, and Min Zhang.
\newblock Neural machine translation advised by statistical machine translation.
\newblock In \emph{Proceedings of the AAAI conference on artificial intelligence}, volume~31, 2017.

\bibitem[xAI(2025)]{grok-3}
xAI.
\newblock Grok 3 beta — the age of reasoning agents, 2025.
\newblock URL \url{https://x.ai/news/grok-3}.
\newblock Accessed: 2025-05-12.

\end{thebibliography}
